\documentclass{article}

\PassOptionsToPackage{numbers}{natbib}
 \usepackage[dblblindworkshop, final]{neurips_2025}
 \usepackage{graphicx} 
 \usepackage{caption}

\usepackage[utf8]{inputenc} 
\usepackage[T1]{fontenc}    
\usepackage{hyperref}       
\usepackage{url}            
\usepackage{booktabs}       
\usepackage{amsfonts}       
\usepackage{nicefrac}       
\usepackage{microtype}      
\usepackage{xcolor}         

\title{FISCAL: Financial Synthetic Claim–document Augmented Learning for Efficient Fact-Checking }
\workshoptitle{Generative AI in Finance}

\author{%
  Rishab Sharma\\
  Charli Capital\\
  Vancouver, Canada\\
  \texttt{rishab@charli.ai} \\
  \And
  Iman Saberi\\
  University of British Columbia\\
  Kelowna, Canada\\
  \texttt{iman@charli.ai} \\
  \And
  Elham Alipour\\
  Charli Capital\\
  Vancouver, Canada\\
  \texttt{elham@charli.ai} \\
  \And
  Jie JW Wu\\
  Michigan Technological University\\
  Houghton, USA\\
  \texttt{jie.jw.wu@mtu.edu} \\
  \And
  Fatemeh Fard \\
  University of British Columbia\\
  Kelowna, Canada\\
  \texttt{fatemeh.fard@ubc.ca} \\
}

\begin{document}

\maketitle

\begin{abstract}
 Financial applications of large language models (LLMs) require factual reliability and computational efficiency, yet current systems often hallucinate details and rely on prohibitively large models. We propose \textsc{FISCAL} (Financial Synthetic Claim–Document Augmented Learning), a modular framework for generating synthetic data tailored to financial fact-checking. We first generate a dataset FISCAL-data using FISCAL. Then we train \textsc{MiniCheck-FISCAL}, a lightweight verifier for numerical financial claims. MiniCheck-FISCAL outperforms its baseline, surpasses GPT-3.5 Turbo and other open-source peers of similar size, and approaches the accuracy of much larger systems (20x) such as Mixtral-8×22B and Command R+. On external datasets FinDVer and Fin-Fact, it rivals GPT-4o and Claude-3.5 while outperforming Gemini-1.5 Flash.
 These results show that domain-specific synthetic data, combined with efficient fine-tuning, enables compact models to achieve state-of-the-art accuracy, robustness, and scalability for practical financial AI. Data and scripts are open-sourced: {\url{https://gitlab.com/charli-ai-public/data-science/fiscal}}. 

\end{abstract}

\section{Introduction}

Large Language Models (LLMs) have transformed the finance domain by enabling applications such as report generation\citep{wu2023bloomberggpt, liu2023fingpt}, question answering\citep{wu2023bloomberggpt, huang2023finbert}, and investment analysis\citep{yang2023investlm}. However, their adoption in production systems remains limited by two critical barriers. First, financial tasks demand \textit{factual reliability}: even small hallucinations in numbers, entities, or dates can lead to costly errors in high-stakes decision-making \citep{prod_in_efficient1, prod_in_efficient2, prod_in_efficient3, liu2025finr1}. Second, state-of-the-art models are computationally inefficient: proprietary systems such as GPT-4 or Gemini are accurate but expensive or slow \citep{memory_inefficient1, memory_inefficient2}. For financial institutions, these constraints make large-scale deployment impractical.

We address these challenges by showing that compact, open models can be both reliable and efficient when equipped with the right training data. Specifically, we introduce \textsc{FISCAL}, a modular data generator that produces diverse and challenging financial claim--document--label triplets, and \textsc{MiniCheck-FISCAL}, a 7B parameter fact-checking model fine-tuned on this data. Unlike prior work that relies on increasingly large LLMs or complex multi-step reasoning pipelines, our approach trains a lightweight verifier that makes fast, single-token predictions with interpretable confidence scores. 

Across three benchmarks, our FISCAL dataset, FinDVer, and Fin-Fact---MiniCheck-FISCAL consistently outperforms its baseline (MiniCheck-7B) and approaches the accuracy of models more than twenty times its size, including,  Mixtral-8x22B, Qwen2-72B, Gemini-1.5-flash, and GPT4o. These results demonstrate that domain-specific synthetic data and a lightweight parameter scale model can yield compact fact-checkers that rival much larger systems, reducing cost during deployment. By reducing inference cost while maintaining high factual accuracy, MiniCheck-FISCAL offers a practical path toward deployable financial AI that is trustworthy, cost-effective, and scalable.

\section{Methodology}

Our approach consists of two main components: (1) a \textbf{Modular Claim--Document Generator (FISCAL)}, which synthesizes claim--document--label triplets (FISCAL-Data) from financial texts, and (2) \textbf{MiniCheck-FISCAL}, a fact-checking model fine-tuned on this synthetic data (FISCAL-Data). Together, these components enable scalable training of domain-adapted fact-checkers while preserving efficiency at inference time.

\subsection{Modular Claim--Document Generator}

The generator takes as input a collection of financial documents $\mathcal{D}=\{d_1,\dots,d_M\}$ and produces a set of labeled triplets $\mathcal{T}=\{(c_i,d_i,y_i)\}_{i=1}^N$, where $c_i$ denotes a claim, $d_i$ the document, and $y_i\in\{1,0\}$ indicates whether the document supports the claim. Constructing $\mathcal{T}$ involves three stages.  

\textbf{Dataset selection.} We begin with FinanceBench \citep{islam2023financebench}, which provides QA pairs with ground-truth evidence. Evidence contexts are extracted to form our base set of documents, ensuring realism and alignment with factual financial data.  

\textbf{Claim extraction.} Because numerical reasoning is a primary source of hallucination in financial LLMs, we focus on the extraction of numerical atomic claims. We employ Qwen3 32B \citep{qwen3} in a prompt-based pipeline to identify quantitative statements. Each extracted claim is verified for atomicity using the LLM-as-judge as explained in section \ref{sec:data-validation}, ensuring that it corresponds to a single, checkable fact.  

\textbf{Data synthesis.} To balance the dataset, we generate both positive and negative examples for each claim--document pair. Rather than relying on a single transformation, we employ a modular synthesizer comprising multiple strategies.  

\begin{itemize}
    \item \textbf{Claim Paraphraser Module.} Generates professionally styled paraphrases of financial claims, preserving all factual details while varying syntax and wording to increase linguistic diversity.

    \item \textbf{Conflict Insertion Module.} Introduces a subtle, contextually consistent contradiction to a financial document by inserting a realistic detail that conflicts with the claim, while leaving all original content otherwise unchanged.
    
    \item \textbf{Fact Exclusion Module.} Removes all evidence of the claim from a financial document, deleting or rewriting only claim-related content while preserving coherence and leaving unrelated text intact.
    
    \item \textbf{Fact Value Distortion Module.} Subtly alters claim-related details in a financial document (e.g., values, dates, entities) with small, plausible distortions, while keeping all unrelated content intact and professionally coherent.
    \item \textbf{Mis-attribution Module.} Edits supporting evidence in a financial document by subtly reassigning attribution (e.g., year, entity, or category), leaving the claim unsupported while preserving coherence and all unrelated content.
    \item \textbf{Summarization Module.} Produces a concise summary of a financial document that retains only claim-relevant details, ensuring the claim remains explicitly verifiable while omitting unrelated content.
\end{itemize}

This modular design ensures coverage of both easy and challenging cases, exposing the model to a broad spectrum of factual errors. In total, we construct 14{,}304 training triplets, 1{,}792 evaluation and 1{,}784 test benchmark from unseen documents in FinanceBench~\citep{islam2023financebench}. 
To prevent data leakage between training and testing, FISCAL-data ensures that no duplicate samples appear across splits.

\subsection{Dataset Validation}\label{sec:data-validation}

To ensure the reliability of the generated dataset, we incorporate multiple rounds of LLM as judges validation.  
During the \textbf{claim extraction stage}, all extracted claims are independently reviewed by two expert LLM-judges (GPT-OSS-120B \cite{gptoss120b} and LLama4 Maverick) for \textit{atomicity}, i.e., whether each claim expresses a single, self-contained factual statement. Only claims with unanimous agreement are retained.  

During the \textbf{data synthesis stage}, two independent LLM-judges validate the correctness of the generated claim--document--label triplets (GPT-OSS-120B \cite{gptoss120b} and LLama4 Scout). Agreement is measured using Cohen’s $\kappa$, yielding a score of 0.892, indicative of strong inter-annotator reliability. To maintain quality and class balance, only triplets with unanimous agreement across all modules are preserved.

\subsection{Fine-Tuning MiniCheck-FISCAL}

We fine-tune \textsc{MiniCheck}-7B \citep{tang2024minicheck}, a lightweight fact-checking model, using parameter-efficient fine-tuning with LoRA \citep{hu2022lora}. Although \textsc{MiniCheck} was originally designed for classification, we reformulate claim verification as a causal language modeling (CLM) task. This enables us to exploit generative capabilities while maintaining efficiency.  
For each triplet $(c_i,d_i,y_i)$, we construct a prompt consisting of a fixed system instruction and the claim--document pair. The model is trained to generate a single-token answer: \texttt{yes} if the document supports the claim ($y_i=1$), or \texttt{no} otherwise ($y_i=0$). The training objective is the negative log-likelihood of the correct token:  
\[
\mathcal{L} = -\log P_\theta(y_i \mid c_i, d_i).
\]

At inference, the model outputs a probability distribution over tokens, defined as follows  
\[
C_i = P_\theta(\texttt{yes} \mid c_i, d_i).
\]  
A threshold $\tau$ is applied to $C_i$ to produce binary predictions. This formulation offers two advantages: (i) inference remains efficient, requiring only a single token probability, and (ii) the resulting confidence scores are directly interpretable, enhancing reliability in high-stakes applications.  

In practice, this setup enables MiniCheck-FISCAL to detect both overt and subtle hallucinations in numerical claims while remaining lightweight enough for deployment in real-world financial systems. By combining modular data generation with an inference-efficient architecture, our approach addresses the dual challenges of factual reliability and computational scalability.

\subsection{Use of LLMs}
All raw prompts were initially authored by the researchers. To enhance their clarity and consistency, the authors employed a large language model (OpenAI’s GPT-5) for linguistic and stylistic refinement. The model’s role was limited strictly to language editing; all conceptual, methodological, and substantive design decisions were made, reviewed, and validated by the research team.

\section{Results}
We evaluate \textbf{MiniCheck-FISCAL} along three key dimensions: (i) benchmark performance, (ii) comparisons to larger LLMs, and (iii) Ablation study, examining contributions of each data module.

\subsection{FISCAL-Data Training Benchmarking}

Table~\ref{table:compare-with-baseline} highlights that MiniCheck-FISCAL(7B) substantially outperforms the baseline MiniCheck-7B across all benchmarks. On the FISCAL dataset, it achieves an F1 of 86.43, with Recall increasing by 26.8 points and Precision by 8.22, showing that domain-aligned training reduces false negatives without sacrificing selectivity.  Among the similar param 7B models MiniCheck-FISCAL outperforms all other models even outperforming at 10x size model Qwen2-72B.

Despite its small size, MiniCheck-FISCAL is \textbf{competitive with 100B\(+\) open-weight models} (e.g., Mixtral-8x22B, Command R+) and \textbf{outperforms proprietary GPT-3.5-Turbo}, while approaching GPT-4o (90.39 F1). This demonstrates that careful supervision can rival brute-force scaling.  

External validation further confirms generalization: on \textbf{FinDVer}, MiniCheck-FISCAL improves F1 by \textbf{+10.84} (70.53 vs.\ 59.69), and on \textbf{Fin-Fact} by \textbf{+7.55} (60.69 vs.\ 53.14), with consistent Recall gains while maintaining high Precision.  

At only 7B parameters, MiniCheck-FISCAL is \textbf{efficient for production deployment}, offering lower cost, faster inference, and easier integration in compliance-sensitive environments compared to massive LLMs or closed APIs.  

Taken together, these results demonstrate that the \textbf{FISCAL dataset equips MiniCheck-FISCAL with both in-domain strength and robust out-of-domain transferability}. The pronounced Recall improvements are especially valuable in financial fact-checking, where missing relevant claims can be as costly as misclassifying them.  

\subsection{Comparison to Larger LLMs}

On the FDV-IE subset of \textbf{FinDVer}, \textsc{MiniCheck-FISCAL} (7B) achieves \textbf{75.6\% accuracy}, as reported in Table~\ref{tab:fdv-ie-rag}. This performance is notable for three reasons. 
First, it substantially outperforms open-source peers of comparable size such as Mistral-7B-v3 (59.5\%), Gemma-7B (59.5\%), and Llama-2-7B (60.0\%). Second, despite being a 7B model, it surpasses much larger open-source systems including Qwen2-72B (68.0\%) and Mixtral-8$\times$22B (70.0\%). Third, it even exceeds proprietary models like Gemini-1.5-Flash (70.5\%), while approaching the performance of frontier LLMs such as GPT-4o (78.5\%) and GPT-3.5-Turbo (79\%).

These findings underscore the effectiveness of FISCAL training: it enables small, efficient models to achieve accuracy levels comparable to state-of-the-art LLMs, while operating at a fraction of the computational cost.

\begin{table*}[t]
\centering
\begin{minipage}[t]{0.50\textwidth}
\centering
\caption{Results on FISCAL-data}
\vspace{2pt}
\renewcommand{\arraystretch}{1.2}
\scalebox{0.62}{
\begin{tabular}{lccccc}
\toprule
\textbf{Model} & \textbf{Dataset} & \textbf{Precision} & \textbf{Recall} & \textbf{F1} & \textbf{Accuracy} \\ 
\midrule
Qwen2-72B &FISCAL-data & \textbf{96.07} & 19.17 & 31.96 & 59.19 \\
Qwen-2.5-7B-instruct &FISCAL-data & 88.04 & 27.24 & 41.61 & 61.77 \\
MiniCheck-7B &FISCAL-data & 79.72 & 58.18 & 67.27 & 71.69 \\ 
\textbf{MiniCheck-FISCAL(7B)} &FISCAL-data & 87.94 & 84.98 & 86.43 & 86.66 \\ 
C4AI Command R+ (104B) &FISCAL-data & 82.61 & 93.72 & 87.82 & 87.00 \\ 
Mixtral-8x22B(141B) &FISCAL-data & 85.85 & 93.83 & 89.66 & 89.18 \\ 
\multicolumn{5}{c}{\textit{Proprietary LLMs}} \\
GPT-3.5-turbo &FISCAL-data & 86.48 & 79.60 & 82.90 & 83.58 \\ 
Claude-3.5-Sonnet &FISCAL-data & 78.93 & 97.87 & 87.39 & 85.87 \\ 
Gemini-1.5-Flash &FISCAL-data & 80.42 & \textbf{98.54} & 88.56 & 87.28 \\ 
GPT-4o &FISCAL-data & 83.48 & \textbf{98.54} & \textbf{90.39} & \textbf{89.52} \\ 
\midrule
\textbf{MiniCheck-FISCAL} &FinDVer & \textbf{89.02} & \textbf{58.40} & \textbf{70.53} & \textbf{75.60} \\ 
MiniCheck-7B &FinDVer & 86.36 & 45.60 & 59.69 & 69.20 \\ 
\midrule
\textbf{MiniCheck-FISCAL} &Fin-Fact & \textbf{87.23} & \textbf{47.43} & \textbf{60.69} & \textbf{62.59} \\ 
MiniCheck-7B &Fin-Fact & 85.49 & 38.55 & 53.14 & 58.61 \\
\bottomrule
\end{tabular}}
\label{table:compare-with-baseline}
\end{minipage}%
\hfill
\begin{minipage}[t]{0.4\textwidth}
\centering
\caption{Accuracy of entailment classification on the FDV-IE subset of FinDVer, using RAG setting.}
\small
\scalebox{0.7}{
\begin{tabular}{l c c}
\toprule
\textbf{Model} & \textbf{\#Params} & \textbf{FDV-IE (RAG)} \\
\multicolumn{3}{c}{\textit{Open-source LLMs}} \\
InternLM2-Math-7b & 7B & 58.5 \\
InternLM2-7B & 7B & 59.5 \\
Gemma-7B & 7B & 59.5 \\
GLM-4-9b & 9B & 61.0 \\
Llama-2-7B & 7B & 60.0 \\
Mistral-7B-v3 & 7B & 59.5 \\
Phi-3-medium-4k & 14B & 61.5 \\
Llama-2-70B & 70B & 61.5 \\
Phi-3-medium-128k & 14B & 61.5 \\
Meta-Llama-3-8B & 8B & 62.5 \\
Yi-1.5-34B & 34B & 62.5 \\
Meta-Llama-3-70B & 70B & 65.5 \\
C4AI Command R+ & 104B & 67.5 \\
Qwen2-72B & 72B & 68.0 \\
Mixtral-8x22B & 141B & 70.0 \\
\textbf{FISCAL-MiniCheck} &\textbf{7B} & \textbf{75.60}\\
\midrule
\multicolumn{3}{c}{\textit{Proprietary LLMs}} \\
Gemini-1.5-Flash & -- & 70.5 \\
GPT-4o & -- & 78.5 \\
GPT-3.5-turbo & -- & 79.0 \\
Claude-3.5-Sonnet & -- & 80.5 \\
\bottomrule
\end{tabular}}
\label{tab:fdv-ie-rag}
\end{minipage}
\end{table*}

\subsection{Ablation Study}
We conducted an ablation study of MiniCheck-FISCAL by removing one augmentation module at a time and evaluating Precision, Recall, F1, and Accuracy (Table \ref{table:ablation-study}, see Appendix~\ref{appendix}). The full model achieves an F1 of 86.43, confirming the need for multiple perturbation types in financial fact-checking. Removing the Claim Paraphraser yields the highest precision (90.44) but collapses recall to 53.03 (F1 66.86), showing its importance for lexical variability; conversely, removing Mis-attribution maximizes recall (88.68) but lowers precision (82.48), indicating its role in filtering permissive matches. Eliminating Summarization sustains high precision (89.97) but reduces recall (79.48), while removing Conflict Insertion lowers precision (80.53) despite stable recall (84.87), highlighting its stabilizing effect. Fact Exclusion and Fact Distortion ablations cause only modest declines (F1 86.08 and 86.28), suggesting redundancy that bolsters robustness. Overall, the results show that recall-oriented modules are indispensable in finance, balanced performance is preferable to optimizing a single metric, and overlapping mechanisms enhance resilience, with the full system’s strength stemming from the synergy of all modules.

\section{Conclusion}

We introduced FISCAL, a modular framework for generating synthetic claim–document pairs for financial fact-checking, and built FISCAL-data, a benchmark dataset derived from this process. Using it, we fine-tuned MiniCheck-FISCAL, a lightweight verifier that integrates targeted claim extraction, augmentation modules, and LoRA fine-tuning to achieve strong factual consistency at low cost.
Experiments show MiniCheck-FISCAL outperforms its baseline and generalizes well to external datasets, reaching accuracy competitive with larger proprietary models like GPT-4o and Claude-3.5. Ablation studies highlight the impact of Conflict Insertion and Claim Paraphraser modules in detecting subtle contradictions and linguistic variation.
Together, FISCAL, FISCAL-data, and MiniCheck-FISCAL demonstrate the value of synthetic, domain-specific data for \textit{compact} and \textit{reliable} fact-checkers in finance. Future work aims for multimodal evidence and results' interpretability.

\bibliographystyle{plainnat}   
\bibliography{aa_ref}

\appendix
\section{Results for Ablation Study of FISCAL Modules}\label{appendix}

\begin{table}[h!]
\centering
\caption{Ablation study of MiniCheck-FISCAL (trained on FISCAL). We remove each module one at a time (one-leave-out) and assess the performance.}
\vspace{2pt}
\renewcommand{\arraystretch}{1.2}
\scalebox{0.85}{
\begin{tabular}{lcccc}
\toprule
\textbf{Variant} & \textbf{Precision} & \textbf{Recall} & \textbf{F1} & \textbf{Accuracy} \\  
\midrule
MiniCheck-FISCAL (full) & 87.94 & 84.98 & \textbf{86.43} & 86.66 \\ 
w/o Claim Paraphraser & 90.44 & 53.03 & 66.86 & 73.71 \\
w/o Conflict Insertion & 80.53 & 84.87 & 82.64 & 82.17 \\
w/o Summarization & 89.97 & 79.48 & 84.40 & 85.31 \\
w/o Mis-attribution & 82.48 & \textbf{88.68} & 85.47 & 84.92 \\
w/o Fact Exclusion & 88.42 & 83.86 & 86.08 & 86.43 \\
w/o Fact Distortion & \textbf{89.22} & 83.52 & 86.28 & \textbf{86.72} \\
\bottomrule
\end{tabular}}
\label{table:ablation-study}
\end{table}

\section{Limitations}
Our study has several limitations. The dataset construction pipeline relies on LLMs as both generators and judges, which may introduce systematic biases relative to human annotators; to mitigate this, we employed multi-stage validation with independent LLM as reviewers and yielding strong inter-annotator agreement (Cohen’s $\kappa=0.892$). We further reduce the model biases by retaining only unanimously agreed triplets. 

Synthetic perturbations cannot capture the full complexity of naturally occurring reporting errors, since real-world financial misstatements often involve subtle narrative shifts, multi-hop inconsistencies, or context-specific omissions that go beyond controlled edits. Nonetheless, our modular design explicitly incorporates multiple perturbation strategies, including paraphrasing, conflict insertion, fact exclusion, value distortion, mis-attribution, and summarization, which together expose the model to diverse linguistic and numerical variations. This diversity helps approximate a broader spectrum of error types, ensuring that the model does not overfit to a single synthetic artifact but instead learns to recognize contradictions, omissions, and distortions across heterogeneous settings. By combining modules that target both surface-level (e.g., paraphrase variation) and deep structural errors (e.g., conflict insertion, mis-attribution), our approach achieves stronger coverage than single method generation pipelines, thereby narrowing (though not eliminating) the gap between synthetic errors and real world reporting pathologies.

Finally, our evaluation frames verification as a binary (\texttt{yes}/\texttt{no}) decision, which simplifies nuanced reasoning involving partial support or multi hop evidence. While this abstraction improves efficiency and interpretability, it omits finer distinctions. We partially address this by reporting interpretable confidence scores, but richer task formulations and evaluation criteria will be necessary for broader deployment.

Overall, \textsc{FISCAL} underscores the promise of modular synthetic data generation for financial verification, yet its effectiveness should be interpreted in light of these limitations.

\section{Example Comparison}

\textbf{Example 1.}

\noindent\textbf{Claim:} The Company issued a \$750,000 promissory note to the CEO on March 29, 2018, transferred to a non-related entity by July 2023, while audit fees of \$45,000 were billed in 2023 before engaging a new auditor. \\

\textbf{Document:} 
certain relationships and related transactions, director independence.
Note Payable, Related Party On March 29, 2018, the Company issued a \$750,000, unsecured promissory note to the Company’s CEO for a potential acquisition and working capital. The note carries an interest rate of 6\% per annum, compounding annually, and matures on December 31, 2022. All principal and interest are due at maturity and there is no prepayment penalty for early repayment of the note. As of September 30, 2023 and 2022, total balance on the debt was \$741,030 and accrued interest totaled \$281,561 and \$223,940, respectively.
In July 2023, this promissory note was purchased by a non-related entity.
On September 30, 2023, the Company owed its Chief Financial Officer \$12,000 for past services to the Company.

PRINCIPAL ACCOUNTANT FEES AND SERVICES Below is the table of audit fees (amounts in US\$) billed by our current auditor, Pinnacle Accountancy Group of Utah (a DBA of Heaton and Company, PLLC) in connection with the audit of our annual financial statements until December 14, 2023 at which time the Company engaged Turner, Stone and Company, LLP to conduct its audit for the year ended September 30, 2023.
| Year Ended September 30, | Audit Services | Audit Related Fees | Tax Fees | Other Fees |
| 2023 | \$ | 37,500 | $ | - | \$ | - | \$ | - |
| 2022 | \$ | 32,500 | $ | - | \$ | - | \$ | - |\\

\textbf{Ground Truth:} The claim is \textbf{not supported}. \\
\textbf{MiniCheck-FISCAL Prediction:} The claim is \textbf{not supported}. \\
\textbf{MiniCheck Prediction:} The claim is \textbf{supported}. \\
\textbf{Dataset:} FinDVer. \\

\textbf{Example 2.}

\noindent\textbf{Claim:} Paysign, Inc.'s market stock price fluctuated between $1.69 and $3.98 in 2023, while raising capital could be dilutive to existing stockholders or impose unfavorable terms, with approximately 32\% of shares controlled by insiders which may block significant changes despite no dividends being paid for the foreseeable future.

\textbf{Document:} 
Additional equity or debt financing may be dilutive to existing stockholders or impose terms that are unfavorable to us or our existing stockholders. We may raise capital in order to provide working capital for our expansion into other products and services using our payments platform. If we raise additional funds by issuing equity securities, our stockholders will experience dilution. Debt financing, if available, may involve arrangements that include covenants limiting or restricting our ability to take specific actions, such as incurring additional debt, making capital expenditures or declaring dividends. Any debt financing or additional equity that we raise may contain terms, such as liquidation and other preferences that are not favorable to us or our current stockholders. If we raise additional funds through collaboration and licensing arrangements with third parties, it may be necessary to relinquish valuable rights to our technologies and products or grant unfavorable license terms.
Risks Related to Ownership of Our Common Stock
Our stock price is volatile and you may not be able to sell your shares at a price higher than what was paid. The market for our common stock is highly volatile. In 2023, our stock price fluctuated between $1.69 and $3.98. The trading price of our common stock could be subject to wide fluctuations in response to, among other things, quarterly variations in operating and financial results, announcements of technological innovations or new products by our competitors or us, changes in prices of our products and services or our competitors’ products and services, changes in product mix, or changes in our revenue and revenue growth rates.
We do not intend to pay dividends for the foreseeable future. We have never declared or paid any cash dividends on our capital stock. We intend to retain any earnings to finance the operation and expansion of our business, and we do not anticipate paying any cash dividends in the foreseeable future. As a result, you will likely receive a return on your investment in our common stock only if the market price of our common stock increases.
Concentration of ownership among our existing directors, executive officers and principal stockholders may prevent new investors from influencing significant corporate decisions. Our directors, executive officers, and holders of more than 5\% of our total shares of common stock outstanding and their respective affiliates, in the aggregate, beneficially own, as of March 22, 2023, approximately 52\% of our outstanding common stock. As a result, these stockholders will be able to exercise a controlling influence over matters requiring stockholder approval, including the election of directors and approval of significant corporate transactions, and will have significant influence over our management and policies for the foreseeable future. Some of these persons or entities may have interests that are different from yours. For example, these stockholders may support proposals and actions with which you may disagree or which are not in your interests. The concentration of ownership could delay or prevent a change in control of our company or otherwise discourage a potential acquirer from attempting to obtain control of our company, which in turn could reduce the price of our common stock. In addition, these stockholders, some of which have representatives sitting on our board of directors (the “Board”), could use their voting control to maintain our existing management and directors in office, delay or prevent changes of control of our company, or support or reject other management and Board proposals that are subject to stockholder approval, such as amendments to our employee stock plans and approvals of significant financing transactions.

\textbf{Ground Truth:} The claim is \textbf{not supported}. \\
\textbf{MiniCheck-FISCAL Prediction:} The claim is \textbf{not supported}. \\
\textbf{MiniCheck Prediction:} The claim is \textbf{supported}. \\
\textbf{Dataset:} FinDVer \\

\textbf{Example 3.}

\noindent\textbf{Claim:} As of December 31, 2023, the company had a cash balance of \$205,718, an accumulated deficit of \$29,922,668, and goodwill on the acquisition of subsidiaries RxCompound and Peaks valued at \$2,164,480, while holding accounts payable of \$515,337 and accrued expenses of \$128,097.

\textbf{Document:} 

Cash and cash equivalents Cash and cash equivalents include all highly liquid debt instruments with original maturities of three months or less which are not securing any corporate obligations. As of December 31, 2023, and March 31, 2023, the Company held a cash balance of $ 205,718 and $ 35,756 , respectively.
Reclassification
No restatement was made in comparative Consolidated Financial Statements. However, certain amounts from the prior year have been reclassified to conform to the current year presentation.
Note 3 — Going Concern The accompanying condensed consolidated financial statements have been prepared assuming that the Company will continue as a going concern. On December 31, 2023, the Company had negative working capital, an accumulated deficit of \$ 29,922,668 . These factors raise substantial doubt about the Company’s ability to continue as a going concern.
SCHEDULE OF GOODWILL

| For the Fiscal Quarter Ended December 31, |
| 2023 | 2022 |
| RxCompound and Peaks | \$ | 2,164,480 | \$ | - |
| Total | \$ | 2,164,480 | \$ | - |
Note 11 — Accounts Payable and Accrued Expenses
Accounts payable and accrued expenses consisted of the following:
Schedule of Accounts Payable and Accrued Expenses

| For the Fiscal Quarter Ended December 31, 2023 |
| Accounts Payable (A) | \$ | 515,337 |
| Accrued Expenses (B) | 128,097 |
| Total | \$ | 643,434 |

\textbf{Ground Truth:} The claim is \textbf{supported}. \\
\textbf{MiniCheck-FISCAL Prediction:} The claim is \textbf{supported}. \\
\textbf{MiniCheck Prediction:} The claim is \textbf{not supported}. \\
\textbf{Dataset:} FinDVer \\

\textbf{Example 4.}

\noindent\textbf{Claim:} Pfizer Inc.'s Q4 2020 short-term borrowings (including current portion of long-term debt of $2,002 million) were $2,703 million.

\textbf{Document:} 

Consolidated Balance Sheets
Pfizer Inc. and Subsidiary Companies
As of December 31,
(MILLIONS, EXCEPT PER COMMON SHARE DATA)
2021
2020
Assets
Cash and cash equivalents
$
1,944 
$
1,786 
Short-term investments
29,125 
10,437 
Trade accounts receivable, less allowance for doubtful accounts: 2021$492; 2020$508
11,479 
7,913 
Inventories
9,059 
8,020 
Current tax assets
4,266 
3,264 
Other current assets
3,820 
3,646 
Total current assets
59,693 
35,067 
Equity-method investments
16,472 
16,856 
Long-term investments
5,054 
3,406 
Property, plant and equipment
14,882 
13,745 
Identifiable intangible assets
25,146 
28,337 
Goodwill
49,208 
49,556 
Noncurrent deferred tax assets and other noncurrent tax assets
3,341 
2,383 
Other noncurrent assets
7,679 
4,879 
Total assets
$
181,476 
$
154,229 
Liabilities and Equity

Short-term borrowings, including current portion of long-term debt: 2021\$1,636; 2020\$2,002
\$
2,241 
\$
2,703 
Trade accounts payable
5,578 
4,283 
Dividends payable
2,249 
2,162 
Income taxes payable
1,266 
1,049 
Accrued compensation and related items
3,332 
3,049 
Deferred revenues
3,067 
1,113 
Other current liabilities
24,939 
11,561 
Total current liabilities
42,671 
25,920 
Long-term debt
36,195 
37,133 
Pension benefit obligations
3,489 
4,766 
Postretirement benefit obligations
235 
645 
Noncurrent deferred tax liabilities
349 
4,063 
Other taxes payable
11,331 
11,560 
Other noncurrent liabilities
9,743 
6,669 
Total liabilities
104,013 
90,756 
Commitments and Contingencies
Preferred stock, no par value, at stated value; 27 shares authorized; no shares issued or outstanding at December 31, 2021 and
December 31, 2020

Common stock, \$0.05 par value; 12,000 shares authorized; issued: 20219,471; 20209,407
473 
470 
Additional paid-in capital
90,591 
88,674 
Treasury stock, shares at cost: 20213,851; 20203,840
(111,361)
(110,988)
Retained earnings
103,394 
90,392 
Accumulated other comprehensive loss
(5,897)
(5,310)
Total Pfizer Inc. shareholders equity
77,201 
63,238 
Equity attributable to noncontrolling interests
262 
235 
Total equity
77,462 
63,473 
Total liabilities and equity
\$
181,476 
\$
154,229 
See Accompanying Notes.
Pfizer Inc.
2021 Form 10-K
53

\textbf{Ground Truth:} The claim is \textbf{supported}. \\
\textbf{MiniCheck-FISCAL Prediction:} The claim is \textbf{supported}. \\
\textbf{MiniCheck Prediction:} The claim is \textbf{not supported}. \\
\textbf{Dataset:} FISCAL-Data \\

\textbf{Example 5.}

\noindent\textbf{Claim:} PepsiCo recorded a tax payment of negative \$309million for the TCJ Act in 2021.   \\

\textbf{Document:} 
Table of Contents
Consolidated Statement of Cash Flows
PepsiCo, Inc. and Subsidiaries
Fiscal years ended December 31, 2022, December 25, 2021 and December 26, 2020
(in millions)
2022
2021
2020
Operating Activities
Net income
\$
8,978 \$
7,679 \$
7,175 
Depreciation and amortization
2,763 
2,710 
2,548 
Gain associated with the Juice Transaction
(3,321)

Impairment and other charges
3,618

Operating lease right-of-use asset amortization
517 
505 
478 
Share-based compensation expense
343 
301 
264 
Restructuring and impairment charges
411 
247 
289 
Cash payments for restructuring charges
(224)
(256)
(255)
Acquisition and divestiture-related charges
80 
(4)
255 
Cash payments for acquisition and divestiture-related charges
(46)
(176)
(131)
Pension and retiree medical plan expenses
419 
123 
408 
Pension and retiree medical plan contributions
(384)
(785)
(562)
Deferred income taxes and other tax charges and credits
(873)
298 
361 
Tax expense related to the TCJ Act
86 
190 
 
Tax payments related to the TCJ Act
(309)
(309)
(78)
Change in assets and liabilities:
Accounts and notes receivable
(1,763)
(651)
(420)
Inventories
(1,142)
(582)
(516)
Prepaid expenses and other current assets
118 
159 
26 
Accounts payable and other current liabilities
1,842 
1,762 
766 
Income taxes payable
57 
30 
(159)
Other, net
(359)
375 
164 
Net Cash Provided by Operating Activities
10,811 
11,616 
10,613 
Investing Activities
Capital spending
(5,207)
(4,625)
(4,240)
Sales of property, plant and equipment
251 
166 
55 
Acquisitions, net of cash acquired, investments in noncontrolled affiliates and purchases of
intangible and other assets
(873)
(61)
(6,372)
Proceeds associated with the Juice Transaction
3,456

Other divestitures, sales of investments in noncontrolled affiliates and other assets
49 
169 
6 
Short-term investments, by original maturity:
More than three months - purchases
(291)
 
(1,135)
More than three months - maturities
150 
1,135 
 
Three months or less, net
24 
(58)
27 
Other investing, net
11 
5 
40 
Net Cash Used for Investing Activities
(2,430)
(3,269)
(11,619)
(Continued on following page)
62\\

\textbf{Ground Truth:} The claim is \textbf{supported}. \\
\textbf{MiniCheck-FISCAL Prediction:} The claim is \textbf{supported}. \\
\textbf{MiniCheck Prediction:} The claim is \textbf{not supported}. \\
\textbf{Dataset:} FISCAL-Data \\

\textbf{Example 6.}

\noindent\textbf{Claim:} The Kraft Heinz Company's Other non-current assets were \$2,100 million as of December 28, 2019. \\
\textbf{Document:} 
The Kraft Heinz Company
Consolidated Balance Sheets
(in millions, except per share data)
 
December 28, 2019 December 29, 2018
ASSETS

Cash and cash equivalents
\$
2,279 \$
1,130
Trade receivables (net of allowances of \$33 at December 28, 2019 and \$24 at December 29, 2018)
1,973 
2,129
Income taxes receivable
173 
152
Inventories
2,721 
2,667
Prepaid expenses
384 
400
Other current assets
445 
1,221
Assets held for sale
122 
1,376
Total current assets
8,097 
9,075
Property, plant and equipment, net
7,055 
7,078
Goodwill
35,546 
36,503
Intangible assets, net
48,652 
49,468
Other non-current assets
2,100 
1,337
Other non-current assets, as detailed in Note 12, were adjusted to \$2,050 million at December 28, 2019, due to reclassification of certain deferred tax items.
TOTAL ASSETS
\$
101,450 \$
103,461
LIABILITIES AND EQUITY

Commercial paper and other short-term debt
\$
6 \$
21
Current portion of long-term debt
1,022 
377
Trade payables
4,003 
4,153
Accrued marketing
647 
722
Interest payable
384 
408
Other current liabilities
1,804 
1,767
Liabilities held for sale
9 
55
Total current liabilities
7,875 
7,503
Long-term debt
28,216 
30,770
Deferred income taxes
11,878 
12,202
Accrued postemployment costs
273 
306
Other non-current liabilities
1,459 
902
TOTAL LIABILITIES
49,701 
51,683
Commitments and Contingencies (Note 17)
 
Redeemable noncontrolling interest
 
3
Equity:

Common stock, \$0.01 par value (5,000 shares authorized; 1,224 shares issued and 1,221 shares outstanding at December 28, 2019;
1,224 shares issued and 1,220 shares outstanding at December 29, 2018)
12 
12
Additional paid-in capital
56,828 
58,723
Retained earnings/(deficit)
(3,060) 
(4,853)
Accumulated other comprehensive income/(losses)
(1,886) 
(1,943)
Treasury stock, at cost (3 shares at December 28, 2019 and 4 shares at December 29, 2018)
(271) 
(282)
Total shareholders' equity
51,623 
51,657
Noncontrolling interest
126 
118
TOTAL EQUITY
51,749 
51,775
TOTAL LIABILITIES AND EQUITY
\$
101,450 \$
103,461
See accompanying notes to the consolidated financial statements.
47\\
\textbf{Ground Truth:} The claim is \textbf{not supported}. \\
\textbf{MiniCheck-FISCAL Prediction:} The claim is \textbf{not supported}. \\
\textbf{MiniCheck Prediction:} The claim is \textbf{supported}. \\
\textbf{Dataset:} FISCAL-Data \\

\textbf{Example 7.}

\noindent\textbf{Claim:} Lockheed Martin's 2020 retained earnings were \$21,636 million.\\

\textbf{Document:} 
Table of Contents
Lockheed Martin Corporation
Consolidated Balance Sheets
(in millions, except par value)

December 31,
2021
2020
Assets
Current assets
Cash and cash equivalents
\$
3,604 
\$
3,160 
Receivables, net
1,963 
1,978 
Contract assets
10,579 
9,545 
Inventories
2,981 
3,545 
Other current assets
688 
1,150 
Total current assets
19,815 
19,378 
Property, plant and equipment, net
7,597 
7,213 
Goodwill
10,813 
10,806 
Intangible assets, net
2,706 
3,012 
Deferred income taxes
2,290 
3,475 
Other noncurrent assets
7,652 
6,826 
Total assets
\$
50,873 
\$
50,710 
Liabilities and equity
Current liabilities
Accounts payable
\$
780 
\$
880 
Salaries, benefits and payroll taxes
3,108 
3,163 
Contract liabilities
8,107 
7,545 
Current maturities of long-term debt
6 
500 
Other current liabilities
1,996 
1,845 
Total current liabilities
13,997 
13,933 
Long-term debt, net
11,670 
11,669 
Accrued pension liabilities
8,319 
12,874 
Other noncurrent liabilities
5,928 
6,196 
Total liabilities
39,914 
44,672 
Stockholders equity
Common stock, \$1 par value per share
271 
279 
Additional paid-in capital
94 
221 
Retained earnings
21,600 
21,636 
However, the subsequent audit adjustment reflected retained earnings of \$21,500 million for the year ended 2020. 
Accumulated other comprehensive loss
(11,006)
(16,121)
Total stockholders equity
10,959 
6,015 
Noncontrolling interests in subsidiary
 
23 
Total equity
10,959 
6,038 
Total liabilities and equity
\$
50,873 
\$
50,710 
The accompanying notes are an integral part of these consolidated financial statements.
68\\

\textbf{Ground Truth:} The claim is \textbf{not supported}. \\
\textbf{MiniCheck-FISCAL Prediction:} The claim is \textbf{not supported}. \\
\textbf{Gemini-Flash-1.5:} The claim is \textbf{supported}. \\
\textbf{Dataset:} FISCAL-Data \\

\end{document}